\documentclass[runningheads,dvipdfmx]{llncs}
\usepackage{graphicx}
\usepackage{booktabs}
\usepackage{subcaption}
\usepackage{mediabb}

\usepackage{hyperref}

\usepackage{orcidlink}

\begin{document}

\title{Spatiotemporal Pooling on Appropriate Topological Maps Represented as Two-Dimensional Images for EEG Classification}

\titlerunning{Spatiotemporal Pooling on Appropriate Topological Maps}

%\author{First Author\inst{1}\orcidlink{0000-1111-2222-3333} \and
%Second Author\inst{2,3}\orcidlink{1111-2222-3333-4444} \and
%Third Author\inst{3}\orcidlink{2222--3333-4444-5555}}
\author{Takuto Fukushima \and Ryusuke Miyamoto\orcidlink{0000-0001-9450-4493}}

\authorrunning{T.~Fukushima and R.~Miyamoto}

%\institute{Princeton University, Princeton NJ 08544, USA \and
%Springer Heidelberg, Tiergartenstr.~17, 69121 Heidelberg, Germany
\institute{Dept. of Computer Science,
  School of Science and Technology,
  Meiji University, Japan\\
%\email{lncs@springer.com}
\email{\{taku,miya\}@cs.meiji.ac.jp}}

\maketitle

\begin{abstract}
  Motor imagery classification
  based on electroencephalography (EEG) signals is one of the most
  important brain-computer interface applications, although it needs
  further improvement. Several methods have
  attempted to obtain useful information from EEG signals by using
  recent deep learning techniques such as transformers. 
  To improve the classification accuracy, this study proposes
  a novel EEG-based motor imagery classification method with three 
  key features: generation of a topological map represented as
  a two-dimensional image from EEG signals with coordinate transformation
  based on t-SNE,
  use of the InternImage to extract spatial features, and use of
  spatiotemporal pooling inspired by PoolFormer to exploit
  spatiotemporal information concealed in a sequence of EEG images. 
  Experimental results using the PhysioNet EEG Motor Movement/Imagery
  dataset showed that the proposed method
  achieved the best classification accuracy of 88.57\%,
  80.65\%, and 70.17\% on two-, three-, and four-class motor imagery
  tasks in cross-individual validation. 
\end{abstract}

\section{Introduction}
\label{sec:intro}
Brain-computer interfaces (BCIs) are expected to enable humans to use
their thoughts as inputs to control various types of devices.
Motor imagery classification is one of the most important BCI applications;
it involves estimating thoughts in a person's brain.
Motor imagery classification can be used for medical applications,
including rehabilitation for stroke patients\cite{lopez2018brain} and 
assistance for individuals unable to walk\cite{6476692},
as well as nonmedical fields, applications,
supporting unmanned vehicles\cite{yu2016toward} and 
personal authentication\cite{zhang2018mindid}. 
To make BCIs practical in such applications,
the motor imagery classification accuracy needs to be
improved\cite{dose2018end}.

Motor imagery classification can be performed using various
approaches such as electroencephalography (EEG),
magnetoencephalography (MEG), and functional magnetic resonance imaging (fMRI).
EEG-based methods have become the most popular because
of their portability, low cost, and high temporal resolution.
In general, estimating long-range temporal correlation
is useful for extracting relevant information from EEG signals
\cite{wairagkar2019modeling,wairagkar2021dynamics} that exhibit
long-range dependencies.
EEG classification has been improved further 
by capturing both temporal and spatial features\cite{altaheri2023deep}.
Recently, models utilizing a transformer\cite{vaswani2017attention}, known for its strength in handling long-range dependencies, 
have been applied for EEG classification and achieved state-of-the-art results\cite{xie2022transformer,du2022eeg,wei2023tc}. 
The effectiveness of a transformer in motor imagery classification tasks
with EEG signals has been demonstrated\cite{xie2022transformer}.

t-CTrans\cite{xie2022transformer} takes an array of raw EEG data as the input,
captures spatial features using a convolutional neural network (CNN), 
adds channel information of the EEG with positional encoding,
and captures temporal correlations using a transformer encoder. 
Its structure enables the effective capture of both spatial and temporal features with state-of-the-art performance. 
However, the accuracy cannot be improved further because
of two problems: 
representation of input data and model structure for temporal
feature extraction.
The first problem is that the input is an array of raw EEG data;
it may not allow the CNN to capture spatial features effectively.
The second problem is that the current implementation adopts
the same structure as that of transformer encoder used in NLP,
as described in \cite{vaswani2017attention},
and it may not extract the temporal correlation of EEG signals appropriately.

To solve the first problem,
we aim to make the spatial features obtained from the input
data more effective. 
The information inputted from EEG signals can be considered
as an image by appropriately projecting the arrangement
of EEG electrodes into a two-dimensional plane;
recent computer vision methods can be applied for this purpose.
This representation called a topological map maintains the
locations of EEG electrodes in two-dimensional coordinates and
keeps the observed values at these coordinates
as pixel values\cite{altaheri2023deep} in a two-dimensional image.

For generating a topological map represented as a two-dimensional
image, the electrode locations
represented by three-dimensional coordinates must be converted appropriately to
two-dimensional ones.
Existing studies used methods such as manual determination\cite{zhao2019multi,liu2021densely}
and azimuthal equidistant projection\cite{bashivan2016learning,sun2021hybrid,9061622} for this purpose.
The manual determination method is problematic because
it requires redesigning when the number of EEG channels varies across
datasets or when the resolution of the topological maps is changed.
Azimuthal equidistant projection transforms coordinates so that distances and angles from the center are accurate; however,
it may lose the relationships between electrodes that are inherently correlated.
Therefore, our proposed method introduces t-distributed stochastic neighbor
embedding (t-SNE)\cite{van2008visualizing}, 
a popular statistical nonlinear dimensionality reduction technique. 
t-SNE has been applied for feature extraction\cite{li2016extracting,ma2021novel}; our study is the first attempt to apply it for coordinate transformation.

By performing appropriate coordinate transformation, 
we can expect to more accurately extract spatial features using computer vision
methods.
This study introduces InternImage\cite{wang2023internimage} to
improve the accuracy of spatial feature extraction beyond
those of conventional methods.
This is because even though InternImage is a CNN-based method,
it achieves higher accuracy than transformer-based methods in certain tasks. 
Additionally, similar to transformer-based methods like Vision Transformer\cite{dosovitskiy2020image} and Swin Transformer\cite{liu2021swin}, 
InternImage can incorporate advancements such as LayerScale
to achieve effective training even as the model structure grows larger.

The second problem is the inadequate handling of temporal information. 
t-CTrans merely use a transformer encoder that is not appropriate approach
when images are arranged in a time series. 
Therefore, we incorporated the concept of PoolFormer\cite{yu2022metaformer},
in which correlations are evaluated
using multi-head attention to adequately handle the structure in which
spatial features are arranged in a time series. 
In contrast, our proposed method treats spatial features as one-dimensional vectors, 
arranges them into a two-dimensional plane corresponding to the spatiotemporal features 
when sequenced over time, and applies 2D pooling to this plane. 
By performing pooling on spatiotemporal features in this way, 
the proposed method is expected to appropriately handle the characteristics of
temporal changes in spatial features.

In summary, the features of the proposed method include appropriately transforming the input format, 
accurately capturing spatial features using InternImage, and 
introducing spatiotemporal pooling (ST-pooling) inspired by PoolFormer to adequately capture the temporal changes in spatial features. 
To evaluate the effectiveness of the proposed method, experiments were
conducted using the PhysioNet EEG Motor Movement/Imagery Dataset\cite{goldberger2000physiobank,bci2000}.
Comparative experiments were performed to assess the coordinate transformation and ST-pooling introduced by the proposed method; this enabled
clarifying the contribution of each component to the improvement in accuracy.

The rest of this paper is organized as follows.
Section~\ref{sec:related} introduces
related work about EEG classification. 
Section~\ref{sec:method} describes a novel method to apply
ST-pooling to topological maps generated from EEG signals. 
Section~\ref{sec:experiment} describes the details of the experimental setup.
Section~\ref{sec:result} presents a comparison of the proposed method with
conventional methods to demonstrate its effectiveness.
Finally, Section~\ref{sec:conc} concludes this paper.

\section{Related Work}
\label{sec:related}
This section describes related research on EEG classification
and explains input representation, coordinate transformation,
and classification model because they are important
when considering input data as images to apply computer vision
techniques.

\subsection{Input representation}
EEG signals are captured as time-series signals from multiple electrodes
simultaneously. To treat these signals as a sequence of
two-dimensional images, they must be reshaped.
Existing image-based analysis methods for transforming EEG signals
using common spatial patterns or
unique methods for feature extraction before model input
\cite{luo2018exploring,olivas2019classification,ma2019deep,zhao2020deep}, 
directly inputting raw data as arrays
\cite{xie2022transformer,du2022eeg,li2019novel,jeong2020eeg}, 
applying transformations like short-time Fourier transform or
continuous wavelet transform to produce spectral images
\cite{xu2018wavelet,tabar2016novel,kant2020cwt,wang2018short}, and 
creating topological maps based on electrode locations\cite{zhao2019multi,liu2021densely,li2019novel,bashivan2016learning}.

However, existing methods have various issues. 
Many of them adopt feature extraction before model input, and they
may not capture relevant information effectively.
Approaches that use positional encoding \cite{xie2022transformer,du2022eeg} or
aligned channels for convolution\cite{li2019novel,jeong2020eeg}
may extract spatial information inappropriately.
Spatial feature extraction remains problematic with
some methods attempt that use spectral images, where different
representation are applied to embed channel locations in a two-dimensional
images, including arranging channels vertically and horizontally in an
image\cite{tabar2016novel,kant2020cwt,wang2018short},
and representing EEG channels as input image channels\cite{xu2018wavelet}.

This study applies computer vision techniques that show
remarkable performance in several deep learning tasks to
topological maps represented as two-dimensional images
and adopt as an input representation, where a pixel value and
the coordinates represent an EEG signal and the electrode location,
respectively.

\subsection{Coordinate transformation}
To represent input data as a sequence of images using topological maps, 
the electrode coordinates in three-dimensional space must be converted
into two-dimensional coordinates. 
Existing methods employ parallel projection or azimuthal
equidistant projection for this conversion. 
Below, the use of t-SNE as well as the abovementioned method for
coordinate transformation is explained.
\begin{figure}[t]
  \centering
  \begin{minipage}{0.8\linewidth}
    \begin{subfigure}{0.49\linewidth}
      \includegraphics[width=\linewidth]{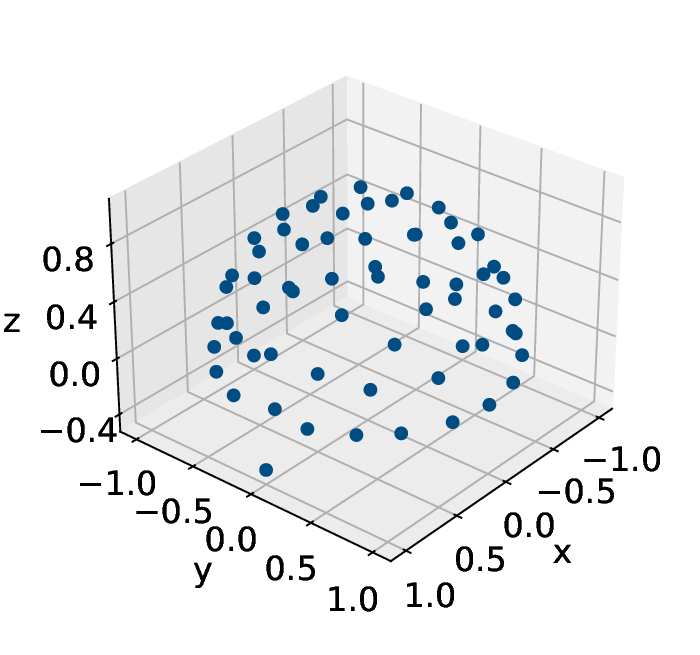}
      \caption{Electrode locations.}
      \label{fig:3d-coordinates}
    \end{subfigure}
    \hfill
    \begin{subfigure}{0.49\linewidth}
        \includegraphics[width=\linewidth]{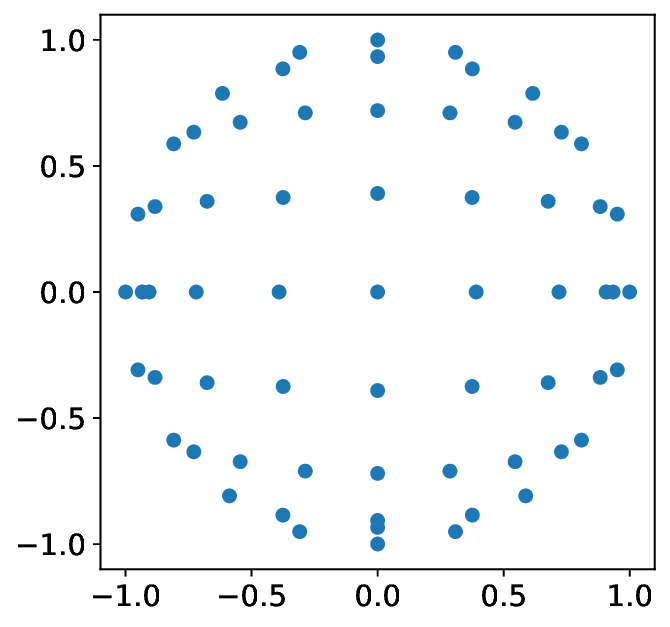}
      \caption{Parallel projection.}
      \label{fig:parallel-projection}
    \end{subfigure}
    \vspace{1cm}
    \begin{subfigure}{0.49\linewidth}
        \includegraphics[width=\linewidth]{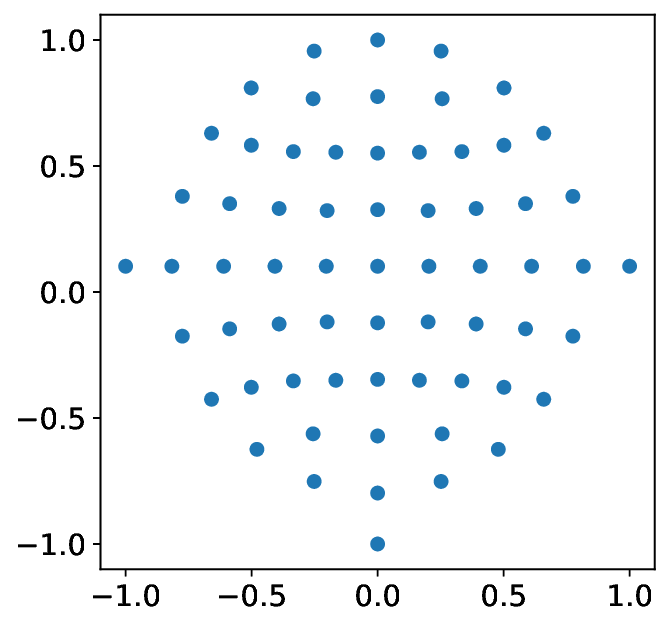}
      \caption{Azimuthal equidistant projection.}
      \label{fig:azim-projection}
    \end{subfigure}
    \hfill
    \begin{subfigure}{0.49\linewidth}
        \includegraphics[width=\linewidth]{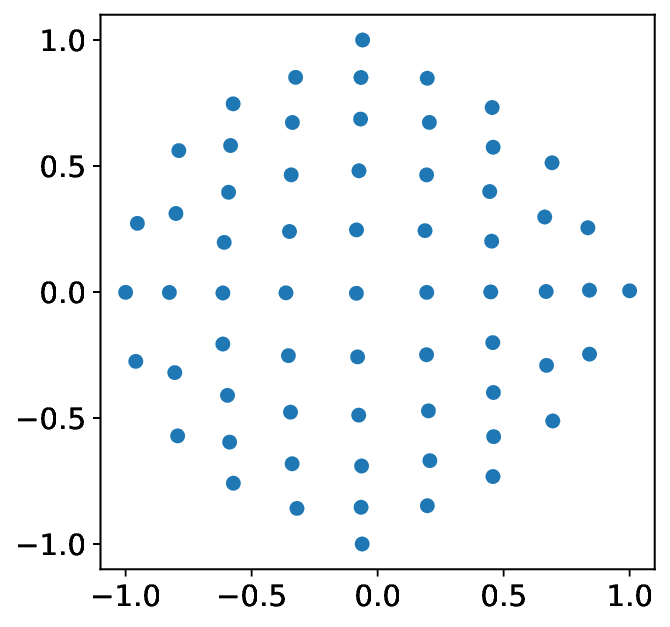}
      \caption{t-SNE.}
      \label{fig:t-SNE}
    \end{subfigure}
    \caption{Three kinds of coordinate transformation that generates two-dimensional coordinates from three-dimensional ones of electrodes.}
    \label{fig:coord_trans}
  \end{minipage}
\end{figure}

First, parallel projection that is the simplest coordinate transformation
is explained. This method involves simplifying the three-dimensional
coordinates of EEG electrodes, 
represented in Euclidean coordinates such as $\left(x,y,z\right)$ by
simply removing the $z-$component to project them onto a two-dimensional plane.
The azimuthal equidistant projection is commonly used for representing the world map, 
notably for depicting shortest flight paths.
It accurately represents
distances and angles from the center point 
but not distances between points far from the center.
t-SNE is a statistical method for visualizing high-dimensional data by performing nonlinear dimensionality reduction. It is designed such that similar data points are placed close to each other in a two-dimensional space, whereas dissimilar data points are positioned further apart.

Figure~\ref{fig:coord_trans} shows the results of transformation to
three-dimensional electrode locations.
Figure~\ref{fig:parallel-projection} shows the result of applying parallel projection, where the $z-$component is simply removed, 
leading to a concentration of electrode coordinates at the ends of the three-dimensional space where the vertical component is significant. 
Figure~\ref{fig:azim-projection} shows the result of applying azimuthal equidistant projection, 
where the distortion increases with distance from the center. Figure~\ref{fig:t-SNE} shows the result of applying t-SNE, where the electrodes are placed more evenly than in other methods.

\subsection{Existing models for motor imagery classification based on EEG signals}
Initially feature extraction methods with a support vector
machine\cite{planelles2014evaluating} 
and k-nearest neighbors \cite{farooq2019motor} were predominantly used
for motor imagery classification. 
However, similar to trends in image classification, improvements in deep learning accuracy have led to the adoption of methods based on deep learning that do not require feature extraction\cite{craik2019deep}. 
These methods utilize EEG data directly or without machine learning as an
input to deep learning models.
For example, there exist models that use only CNN\cite{lawhern2018eegnet,amin2019multilevel}, only long short-term memory (LSTMs)\cite{wang2018lstm}, combinations of CNN and LSTM\cite{zhu2019study,yang2018deep} etc.
CNN-based methods are widely used for motor imagery classification;
however they have limitations in adequately handling long-distance dependencies,
and consequently, 
their accuracy can be improved to a limited extent\cite{craik2019deep,song2021transformer}.
Thus, similar to trends in NLP and general computer vision tasks, 
studies have started using transformer\cite{vaswani2017attention} for
motor imagery classification 
following the introduction of the attention mechanism\cite{xie2022transformer}. 
However, methods focusing on the correlation of spatial features over time
remain lacking, and their accuracy can be improved further.

\section{ST-pooling with Appropriate Topological Maps}
\label{sec:method}
This study proposes a method for motor imagery classification
based on EEG signals with three key features to
improve the classification accuracy: generation of a topological map
represented as a two-dimensional image from EEG signals, extraction
of spatial features with InternImage, and ST-pooling
inspired by PoolFormer\cite{yu2022metaformer}.

\subsection{Generation of a topological map}
The primary feature of the proposed method is the generation of a topological map 
represented as a two-dimensional image from EEG signals. 
For this purpose,
electrode locations that are originally expressed in three-dimensional
coordinates must be appropriately represented in two-dimensional coordinates. 
Among the various methods described in Sec.~\ref{sec:related},
dimensionality reduction using t-SNE is adopted for coordinate transformation.
It needs initial values to perform iterative computations.
For determining the initial values, similar to the implementation in scikit-learn, 
the results of independent component analysis (ICA)
on the set of three-dimensional electrode coordinates are used. 
However, whereas the original implementation intended for application to 
large datasets and therefore calculates singular values by performing
singular value decomposition (SVD)
on randomly selected samples during the ICA process, the implementation in this study 
performs SVD on all samples. 
This is because the number of samples is small, 
and acceleration through random sampling is unnecessary. 
Among the obtained singular values, the two largest ones are selected.

\begin{figure}[tb]
  \centering
  \includegraphics[width=\linewidth]{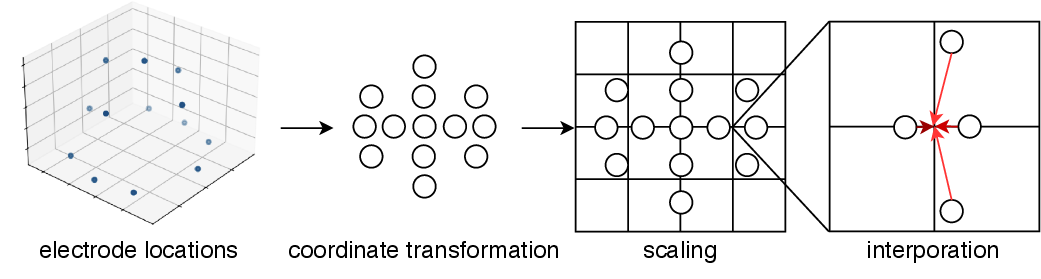}
  \caption{Generation of a topological map.}
  \label{fig:topological-flow}
\end{figure}

By applying t-SNE, the two-dimensional coordinates necessary for
generating a topological map from the three-dimensional coordinates
of the electrodes are obtained. 
These two-dimensional coordinates 
are not integer values.
In addition, scaling is necessary to match the size of the input image 
to the subsequent processes. To obtain appropriate representations
using two-dimensional images whose coordinates are integer values,
interpolation similar to image interpolation is necessary:
the nearest neighbor method was adopted because
it showed the best accuracy in preliminary experiments. 
Finally, a topological map represented as a two-dimensional image 
from EEG signals is obtained:
the number of channels is one, and the height $H$ and width $W$can be
determined arbitrarily by scaling.
This process is shown in Fig.~\ref{fig:topological-flow}.

\subsection{ST-pooling for EEG signals classification}
\begin{figure}[tb]
  \centering
  \includegraphics[width=\linewidth]{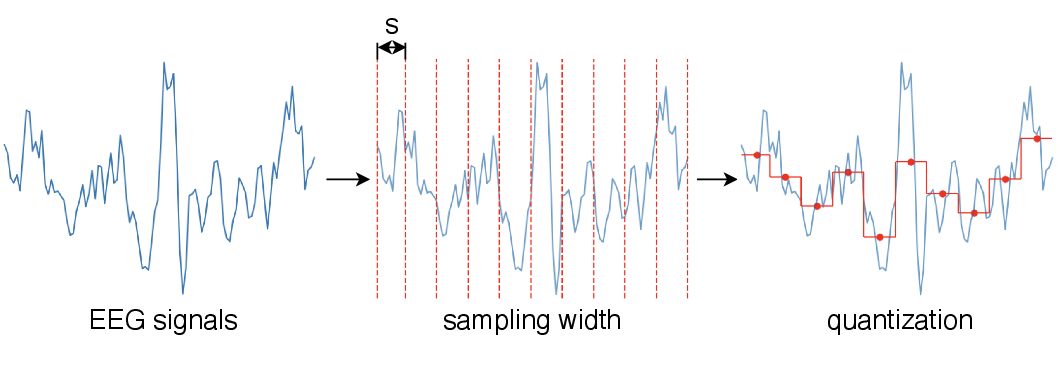}
  \caption{Quantization process of EEG signals.}
  \label{fig:quantization}
\end{figure}

By using the above process, 
EEG signals at a given moment can be treated as two-dimensional images. 
In this study, we propose a new model aimed at 
capturing the temporal changes in EEG signals
with appropriate time-series images as inputs. 
However, considering the generally high sampling frequency of EEG signals, 
directly converting these signals into input images is not 
practical owing to the limitation of computational resources.
Therefore, as shown in Fig.~\ref{fig:quantization}, 
quantization along the time direction is performed. 
In this process, 
the signals at each electrode are averaged over a fixed number of samples $s$, 
and this average is taken as the representative value for that interval.

\begin{figure}[tb]
  \centering
  \includegraphics[width=\linewidth]{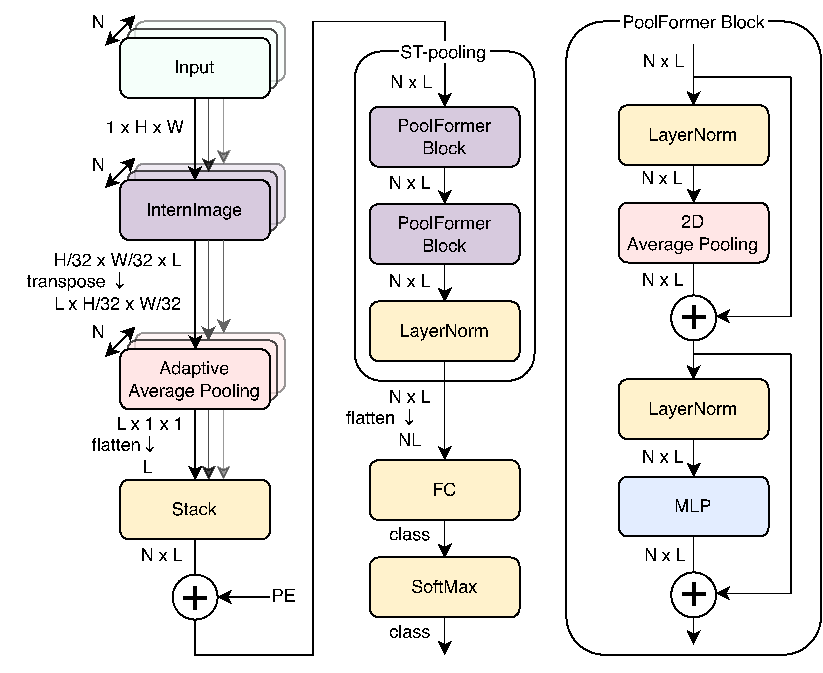}
  \caption{
   Architecture of proposed model with ST-pooling. $N$, $H$, $W$,
    and $L$ means the number of input frames, height of topological maps,
    width of topological maps, and length of a feature vector from InternImage, respectively.
  }
  \label{fig:model}
\end{figure}

We apply the proposed model to the input spatiotemporal images generated in this manner as, shown in Fig.~\ref{fig:model}. 
As shown on the left side of Fig.~\ref{fig:model}, 
the spatial features of each input frame are extracted through an InternImage process. 
However, to obtain the feature vectors used for subsequent ST-pooling, 
adaptive average pooling is applied followed by conversion
into a one-dimensional format. 
This gives the following
feature vector having length $L$:
\begin{equation}
  L = 2^{stage-1} \times C_1,
  \label{eq:L}
\end{equation}
where $stage$ and $C_1$ are parameters depending on the size of
InternImage.

Positional encoding is performed to retain the temporal correlations of the feature vectors obtained from InternImage
during the learning process.
The widely used positional encoding in transformer-based methods, 
expressed by the following equation\cite{vaswani2017attention}, is applied:
\begin{eqnarray}
  PE(n, 2l)   &=& \sin \left(\frac{frame}{10000^{2l/L}}\right), and\\
  PE(n, 2l+1) &=& \cos \left(\frac{frame}{10000^{2l/L}}\right),
\end{eqnarray}
where $n$ and $l$ are the frame number of an input sample
and index of a one-dimensional feature vector, respectively.

Next, to extract spatiotemporal features, 
ST-pooling inspired by PoolFormer is performed. 
In this process, the spatial features outputted from InternImage are 
treated as one-dimensional vectors. 
These vectors are then arranged in chronological order for  $N$ input frames,
resulting in a two-dimensional plane. 
The aim of this process is to extract spatio-temporal features by applying 2D pooling instead of aplying multi-head attention applied as in previous
tTransformer-based studies.

A previous study reported that 
Post-Norm leads to difficulties in training as models become deeper\cite{wang2019learning}; therefore, the proposed method, 
which employs a deep model structure in conjunction with InternImage, adopts Pre-Norm.
As suggested in previous studies,
to maintain the norm scale of the final output, 
a LayerNorm layer was added just before the final output. 
In addition to Pre-Norm, to address the challenges 
posed by the deep model structure, the LayerScale method 
that facilitates efficient learning in deep models, was introduced.

The flow of the proposed method with these processes
is shown in Fig.~\ref{fig:model}. The input through the LayerNorm layer
and then through a fully connected (FC) layer and the SoftMax function, 
resulting in the final output.

\section{Experimental setup}
\label{sec:experiment}
This section describes the dataset and the
approach to train the classifiers used in the evaluation.

\subsection{Dataset}
This study aims to minimize the user-dependent calibration
incurred in constructing BCIs.
Therefore, instead of within-individual training, 
evaluations is conducted through cross-individual training, 
where samples obtained from different subjects are mixed
into training and testing sets.

PhysioNet EEG Motor Movement/Imagery Dataset and 
BCI Competition IV 2a\cite{brunner2008bci} and 2b\cite{leeb2008bci}
are commonly used datasets for motor imagery classification.
However, BCI Competition IV 2a and 2b have only nine subjects;
this is too few samples for evaluating cross-individual training. 
Therefore, the PhysioNet EEG Motor Movement/Imagery dataset was adopted.

This dataset
includes tasks involving motor imagery and actual movement. 
The EEG sampling frequency is 160Hz, 
and the electrode placement follows the 10-10 system. 
The number of subjects is 109, 
with each subject performing 14 trials in a single task session. 
The tasks are as follows:
\begin{itemize}
  \item Baseline, eyes open (O)
  \item Three task runs for motor imagery of left fist (L)
  \item Three task runs for motor imagery of right fist (R)
  \item Three task runs for motor imagery of both feet (F)
\end{itemize}
Based on these labels as in 
previous studies\cite{dose2018end,xie2022transformer}, 
the tasks were divided into three categories for accuracy measurement: 
four-class (L/R/O/F), three-class (L/R/O), and two-class (L/R) classification.
However, owing to inaccuracies in the annotation data, 
the data from only 103 subjects, 
excluding subjects no. 38, 88, 89, 92, 100, and 104, 
were used in the experiments\cite{roots2020fusion}.

\subsection{Cross-individual training}
Subjects were randomly divided into five blocks, 
with one block serving as the test data and the remaining four blocks 
used as training and validation data. 
The accuracy of the test data for all five blocks was determined, 
and their average was taken as the final accuracy. 
Off all subjects, 87.5\% (i.e. 70\% of total number of subjects)
were randomly assigned to the training data, and
the remaining subjects allocated to the validation data. 
The model that achieved the highest accuracy on the validation data was 
used for the test. 
This protocol is a standard one
that has been used previously\cite{roots2020fusion} except for model selection.

\subsection{Data augmentation}

Data augmentation is crucial for obtaining a good model. 
In this study, commonly used deep learning techniques
such as MixUp \cite{zhang2017mixup} and CutMix \cite{yun2019cutmix} 
were applied. 
Furthermore, 
augmentation was performed by adding Gaussian noise to the EEG data used 
for creating topological maps.
This is because noise can be introduced during the measurement of EEG signals. The noise used here has a mean of 0 and a standard deviation of 1, scaled down by $1e-4$.

RandAugment \cite{cubuk2020randaugment} was not adopted though it is
known to be effective in image recognition tasks, was not adopted. 
This is because the data targeted in this study has 
the same format as images, 
in which values are arranged on a two-dimensional plane;
however, the positional information of up, down, left, and right is 
shown as very important in motor imagery
tasks \cite{ehrsson2003imagery}, 
and it was considered necessary to preserve this positional information.

\subsection{Training conditions}
\begin{table}
  \caption{Parameters for training.}
  \label{tab:setup}
  \centering
  \begin{tabular}{p{0.38\linewidth}c}
    \toprule
    parameters & value\\
    \midrule
    learning rate  & $1e-4$ \\
    batch size & 8\\
    drop rate (InternImage) & 0.0\\
    drop rate (ST-pooling) & 0.1\\
    drop path rate (InternImage) & 0.4\\
    drop path rate (ST-pooling) & 0.0\\
  \bottomrule
  \end{tabular}
\end{table}

FractalDB \cite{kataoka2020pre} was used for the pre-training of InternImage.
The rationale behind this choice was that pre-training with FractalDB 
achieved higher accuracy than pre-training with ImageNet \cite{ILSVRC15} 
on some image classification datasets. 
Additionally, because FractalDB consists of artificial images rather than 
natural images, it was deemed suitable 
for the pre-training of topological maps like those used in this study. 
The batch size was set to 256, 
and the learning rate was set at 0.0005 for the pre-training process. 
Based on the obtained model's accuracy, 
the three- and four-class tasks were trained for 40 epochs, and
the two-class task was trained for 30 epochs.

The training parameters used 
when training inference models with the targeted dataset are 
shown in Tab.~\ref{tab:setup}.
In our current implementation, the number of InternImage models
included in the proposed method equals the number of frames
in an input sample.
As a result, a small batch size was used considering the limitation
of computational resources, even though InternImage-S, one of the smallest
implementations of InternImage, with
$stage$ and $C_1$ in Eq.~(\ref{eq:L}) as 4 and 80, respectively.
Adam \cite{kingma2014adam} was used as the optimizer and 
training was conducted for 50 epochs.

\section{Results and discussion}
\label{sec:result}
This section presents the experimental results and discusses
about the effect of the components
implemented in the proposed architecture.

\subsection{Preliminary experiments}
The proposed model includes several parameters to be defined before
the quantitative evaluation. These parameters are determined by
considering the results of the preliminary experiments described below.

First, the number of frames $N$ for an input sample whose length was six seconds
was determined. Table~\ref{tab:comp_timepoint} shows the classification accuracy
of the four-class motor imagery task in cross-individual validation.
\begin{table}
  \begin{minipage}{.3\linewidth}
    \centering
    \caption{Classification accuracy for several frame sizes.}
    \label{tab:comp_timepoint}
    \begin{tabular}{lc}
      \toprule
      t & L/R/O/F \\
      \midrule
      30 & 69.61 \\
      60 & \bf{70.17} \\
      96 & 69.60  \\
      120 & 68.41  \\
      \bottomrule
    \end{tabular}
  \end{minipage}
  \begin{minipage}{.3\linewidth}
  \centering
  \vspace{-.4cm}
  \caption{Classification accuracy when different datasets were used for pre-training.}
  \label{tab:pretrain-comp}
  \begin{tabular}{lc}
    \toprule
    Dataset & L/R/O/F \\
    \midrule
    FractalDB-1k & \bf{70.17}\\
    ImageNet-1k & 69.16\\
    \bottomrule
  \end{tabular}
  \end{minipage}
  \begin{minipage}{.3\linewidth}
    \centering
    \vspace{-.4cm}
    \caption{Classification accuracy when the interpolation method was changed.}
    \label{tab:interpolation-comp}
    \begin{tabular}{lc}
      \toprule
      Method & L/R/O/F \\
      \midrule
      nearest neighbor & \bf{70.17} \\
      Clough-Tocher \cite{ALFELD1984169} & 69.29 \\
      \bottomrule
    \end{tabular}
  \end{minipage}
\end{table}
Considering these results, $N$ was set to 60.

Next, we compare the pre-training datasets: FractalDB and ImageNet.
Table~\ref{tab:pretrain-comp} shows the classification accuracy of
the four-class motor imagery task in cross-individual validation.
Considering these results, Fractal DB was adopted for pre-training.

Finally, the interpolation method used in the generation of topological maps
was chosen.
Table~\ref{tab:interpolation-comp} shows the classification accuracy of
the four-class motor imagery task in cross-individual validation
when the interpolation method was changed.
The results showed that simple interpolation based on the
nearest neighbor method was better than the sophisticated method
widely used in EEG signals classification.

\subsection{Experimental results}
Table~\ref{tab:result} shows the classification of all tasks by the proposed
model, t-CTrans, CNN, and EEGNet Fusion.
The classification accuracy achieved by the proposed model reached
88.57\%, 80.65\%, and 70.17\% on two-, three-, and four-class
motor imagery tasks,
respectively, in cross-individual validation.
These values are better than those of all other state-of-the-art methods
for this task. 

\begin{table}
  \centering
  \caption{Classification accuracy (\%) in the PhysioNet Dataset in cross-individual classification. Only EEGNet Fusion used data whose length was 4s.}
  \label{tab:result}
  \begin{tabular}{p{4cm}|p{2.5cm}p{2.5cm}p{2.5cm}}
    \toprule
    Models & \multicolumn{1}{c}{L/R} & \multicolumn{1}{c}{L/R/O} & \multicolumn{1}{c}{L/R/O/F} \\
    \midrule
    ours & \multicolumn{1}{c}{\bf{88.57}} & \multicolumn{1}{c}{\bf{80.65}} & \multicolumn{1}{c}{\bf{70.17}} \\
    t-CTrans(2022)\cite{xie2022transformer} & \multicolumn{1}{c}{87.80} & \multicolumn{1}{c}{78.98} & \multicolumn{1}{c}{68.54} \\
    CNN(2018)\cite{dose2018end} & \multicolumn{1}{c}{87.98} & \multicolumn{1}{c}{76.61} & \multicolumn{1}{c}{65.73} \\
    EEGNet Fusion(2020)\cite{kostas2021bendr} & \multicolumn{1}{c}{83.80} & \multicolumn{1}{c}{-} & \multicolumn{1}{c}{-} \\
    ConTraNet(2022)\cite{ali2022contranet} & \multicolumn{1}{c}{83.61} & \multicolumn{1}{c}{74.38} & \multicolumn{1}{c}{65.44} \\
    \bottomrule
  \end{tabular}
\end{table}

\begin{figure}[tb]
  \centering
  \includegraphics[width=0.7\linewidth]{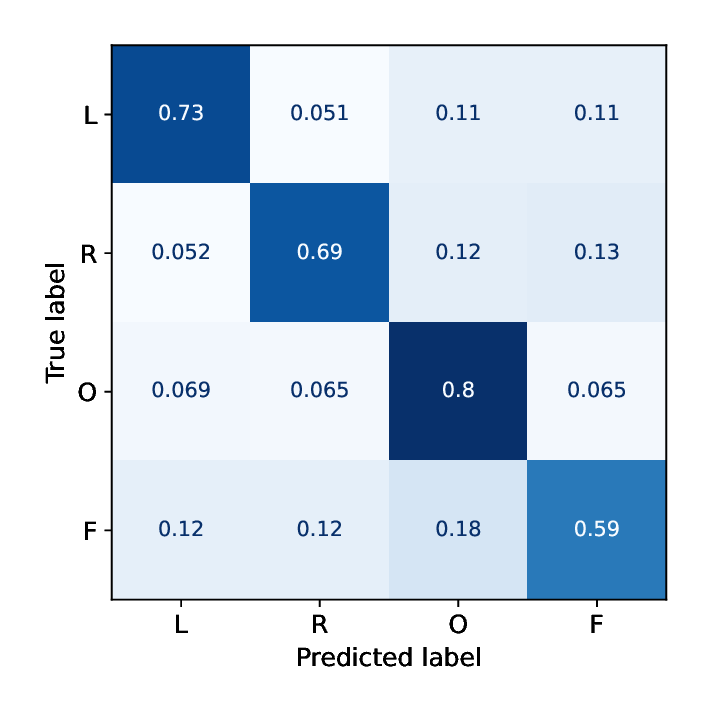}
  \caption{Confusion matrix of four-class classification task.}
  \label{fig:4class_confusion}
\end{figure}

Next, the classification results were analyzed by considering 
the confusion matrix for the four-class task shown 
in Fig.~\ref{fig:4class_confusion}.  
The accuracy was 0.73\%, 0.69\%, 0.80\%, and 0.59\% for the estimation of
L, R, O, and F, respectively.
The accuracy values for L, R, and O were similar, and that for F became
was slightly lower.
This was attributed to the same cause reported previously
\cite{ehrsson2003imagery,10.1093/gigascience/gix034}:
EEG signals for the estimation of task F do not
have significant differences from those for other tasks, although
their intensity increased for tasks O
and changed drastically for tasks L and R.
To further improve the performance of this task,
we must construct a classifier that can extract
hidden features in the current representation of input signals.

\subsection{Effect of proposed components}

\begin{table}
  \begin{minipage}{.45\linewidth}
    \centering
    \caption{Classification accuracy when topological maps generation method was changed.}
    \label{tab:coordinate-comp}
    \begin{tabular}{lc}
      \toprule
      Method & L/R/O/F \\
      \midrule
      parallel projection & 69.30 \\
      azimuthal equidistant & 68.71 \\
      t-SNE & \bf{70.17} \\
      UMAP\cite{mcinnes2018umap} & 68.42 \\
      \bottomrule
    \end{tabular}
    \vspace{-.5cm}
  \end{minipage}
  \begin{minipage}{.45\linewidth}
    \centering
    \vspace{-.3cm}
    \caption{Classification accuracy when the model structure was changed.}
    \label{tab:model-comp}
  \begin{tabular}{lc}
    \toprule
    Method & L/R/O/F \\
    \midrule
    ST-pooling & \bf{70.17}\\
    Multi-Head Attention & 69.88\\
    PoolFormer\cite{yu2022metaformer} & 69.74 \\
    \bottomrule
  \end{tabular}
  \end{minipage}
\end{table}

To analyze the effect of using the proposed method to generate topological maps
from EEG signals, the classification accuracy achieved using the
different methods is shown in Tab.~\ref{tab:coordinate-comp}.
When using the method based on t-SNE, 
the accuracy improved by 0.87\% and 1.46\% compared
to those of simple parallel projection and azimuthal equidistant projection,
respectively.
The results showed that t-SNE is more accurate than UMAP, 
another nonlinear dimensionality reduction method, 
and that t-SNE is suitable 
for generating topological maps for motor imagery classification 
from EEG signals.

To verify the effect of ST-pooling,
a quantitative comparison with the multi-head attention
used in existing methods and PoolFormer was conducted:
the ST-pooling block was replaced with multi-head attention
and PoolFormer takes spatiotemporal images after InternImage
and the stacking process in the same way as in the proposed method.
Table~\ref{tab:model-comp} shows the classification accuracy.

Replacing multi-head attention
with an ST-pooling layer resulted in an accuracy improvement of 0.29\%.
The accuracy with ST-pooling was better than PoolFormer by 0.43\%
even though ST-pooling was inspired by PoolFormer\cite{yu2022metaformer}.
These results showed that ST-pooling successfully captured
the appropriate spatiotemporal correlations
concealed in topological maps for an input sample
generated from successive EEG signals for this task.

\section{Conclusion}
\label{sec:conc}
This study proposed a novel method for motor imagery classification
based on EEG signals with three key features for improving accuracy:
topological map generation with appropriate coordinate transform,
use of InternImage to extract spatial features, and
use of ST-pooling to treat spatial and temporal features
concealed in a sequence of EEG images.
To realize appropriate coordinate transformation,
t-SNE, which was originally proposed for nonlinear dimensionality
reduction was adopted instead of other simpler methods and
other dimensionality reduction methods.
ST-pooling was designed by replacing the multi-head attention
in PoolFormer to exploit the spatiotemporal information concealed
in a sequence of EEG images.

Experimental results using the PhysioNet EEG Motor Movement/Imagery dataset
showed that the proposed method improved the classification accuracy to
88.57\%, 80.65\%, and 70.17\% on two-, three-, and four-class motor imagery
tasks, respectively,
in cross-individual validation; all of these are better than results
obtained using existing state-of-the-art methods.
Further, an analysis using the four-class
task showed the effectiveness of the components included in the proposed
method: topological map generation based on t-SNE improved the accuracy
by 1.46\% compared with that of azimuthal equidistant projection, and
ST-pooling improved the accuracy by 0.29\% compared with that of
multi-head attention. Overall,
when spatial feature extraction using InternImage was adopted in addition to
these components,
the classification accuracy was improved
by 1.63\% compared with that of t-CTrans\cite{xie2022transformer},
which showed the best accuracy among existing methods.

This study showed that the proposed method's functions
to treat spatiotemporal features in a sequence of EEG signals
represented using topological maps were effective for motor imagery
classification. To further demonstrate the capability of the proposed method,
we will apply it to other tasks
using EEG signals and improve its performance
in the future.

\par\vfill\par
% \clearpage  % TODO REVIEW/FINAL: This \clearpage needs to be removed from both review and camera-ready versions.

\bibliographystyle{splncs04}
\bibliography{main}

\begin{thebibliography}{10}
\providecommand{\url}[1]{\texttt{#1}}
\providecommand{\urlprefix}{URL }
\providecommand{\doi}[1]{https://doi.org/#1}

\bibitem{ALFELD1984169}
Alfeld, P.: {A trivariate {C}lough―{T}ocher scheme for tetrahedral data}.
  Comput. Aided Geom. Des.  \textbf{1}(2),  169--181 (1984).
  \doi{https://doi.org/10.1016/0167-8396(84)90029-3}

\bibitem{ali2022contranet}
Ali, O., Saif-ur Rehman, M., Glasmachers, T., Iossifidis, I., Klaes, C.:
  {ConTraNet: A single end-to-end hybrid network for EEG-based and EMG-based
  human machine interfaces}. arXiv preprint arXiv:2206.10677  (2022).
  \doi{10.48550/arXiv.2206.10677}

\bibitem{altaheri2023deep}
Altaheri, H., Muhammad, G., Alsulaiman, M., Amin, S.U., Altuwaijri, G.A.,
  Abdul, W., Bencherif, M.A., Faisal, M.: {Deep learning techniques for
  classification of electroencephalogram (EEG) motor imagery (MI) signals: a
  review}. Neural Comput. Appl.  \textbf{35}(20),  14681--14722 (2023).
  \doi{10.1007/s00521-021-06352-5}

\bibitem{amin2019multilevel}
Amin, S.U., Alsulaiman, M., Muhammad, G., Bencherif, M.A., Hossain, M.S.:
  {Multilevel Weighted Feature Fusion Using Convolutional Neural Networks for
  EEG Motor Imagery Classification}. IEEE Access  \textbf{7},  18940--18950
  (2019). \doi{10.1109/ACCESS.2019.2895688}

\bibitem{bashivan2016learning}
Bashivan, P., Rish, I., Yeasin, M., Codella, N.: {Learning Representations from
  EEG with Deep Recurrent-Convolutional Neural Networks}. In: ICLR. pp. 1--15
  (2016)

\bibitem{brunner2008bci}
Brunner, C., Leeb, R., M{\"u}ller-Putz, G., Schl{\"o}gl, A., Pfurtscheller, G.:
  {BCI Competition 2008--Graz data set A}. Institute for Knowledge Discovery
  (Laboratory of Brain-Computer Interfaces), Graz University of Technology
  \textbf{16}, ~1--6 (2008)

\bibitem{6476692}
Carlson, T., del R.~Millan, J.: {Brain-Controlled Wheelchairs: A Robotic
  Architecture}. IEEE Robot. Autom. Mag.  \textbf{20}(1),  65--73 (2013).
  \doi{10.1109/MRA.2012.2229936}

\bibitem{10.1093/gigascience/gix034}
Cho, H., Ahn, M., Ahn, S., Kwon, M., Jun, S.C.: {EEG datasets for motor imagery
  brain-computer interface}. GigaScience  \textbf{6}(7),  gix034 (2017).
  \doi{10.1093/gigascience/gix034}

\bibitem{craik2019deep}
Craik, A., He, Y., Contreras-Vidal, J.L.: {Deep learning for
  electroencephalogram (EEG) classification tasks: a review}. J. Neural Eng.
  \textbf{16}(3),  031001 (2019). \doi{10.1088/1741-2552/ab0ab5}

\bibitem{cubuk2020randaugment}
Cubuk, E.D., Zoph, B., Shlens, J., Le, Q.V.: {Randaugment: Practical automated
  data augmentation with a reduced search space}. In: CVPR. pp. 702--703 (2020)

\bibitem{dose2018end}
Dose, H., M{\o}ller, J.S., Iversen, H.K., Puthusserypady, S.: {An end-to-end
  deep learning approach to MI-EEG signal classification for BCIs}. Expert
  Syst. Appl.  \textbf{114},  532--542 (2018). \doi{10.1016/j.eswa.2018.08.031}

\bibitem{dosovitskiy2020image}
Dosovitskiy, A., Beyer, L., Kolesnikov, A., Weissenborn, D., Zhai, X.,
  Unterthiner, T., Dehghani, M., Minderer, M., Heigold, G., Gelly, S.,
  Uszkoreit, J., Houlsby, N.: {An Image is Worth 16x16 Words: Transformers for
  Image Recognition at Scale}. In: ICLR (2021)

\bibitem{du2022eeg}
Du, Y., Xu, Y., Wang, X., Liu, L., Ma, P.: {EEG temporal--spatial transformer
  for person identification}. Sci. Rep.  \textbf{12}(1),  14378 (2022).
  \doi{10.1038/s41598-022-18502-3}

\bibitem{ehrsson2003imagery}
Ehrsson, H.H., Geyer, S., Naito, E.: {Imagery of Voluntary Movement of Fingers,
  Toes, and Tongue Activates Corresponding Body-Part-Specific Motor
  Representations}. J. Neurophysiol.  \textbf{90}(5),  3304--3316 (2003).
  \doi{10.1152/jn.01113.2002}

\bibitem{9061622}
Fadel, W., Kollod, C., Wahdow, M., Ibrahim, Y., Ulbert, I.: {Multi-Class
  Classification of Motor Imagery EEG Signals Using Image-Based Deep Recurrent
  Convolutional Neural Network}. In: BCI. pp.~1--4 (2020).
  \doi{10.1109/BCI48061.2020.9061622}

\bibitem{farooq2019motor}
Farooq, F., Rashid, N., Farooq, A., Ahmed, M., Zeb, A., Iqbal, J.: {Motor
  Imagery based Multivariate EEG Signal Classification for Brain Controlled
  Interface Applications}. In: ICOM. pp.~1--6. IEEE (2019).
  \doi{10.1109/ICOM47790.2019.8952008}

\bibitem{goldberger2000physiobank}
Goldberger, A.L., Amaral, L.A., Glass, L., Hausdorff, J.M., Ivanov, P.C., Mark,
  R.G., Mietus, J.E., Moody, G.B., Peng, C.K., Stanley, H.E.: {PhysioBank,
  PhysioToolkit, and PhysioNet Components of a New Research Resource for
  Complex Physiologic Signals}. Circulation  \textbf{101}(23),  e215--e220
  (2000). \doi{10.1161/01.CIR.101.23.e215}

\bibitem{jeong2020eeg}
Jeong, J.H., Lee, B.H., Lee, D.H., Yun, Y.D., Lee, S.W.: {EEG Classification of
  Forearm Movement Imagery Using a Hierarchical Flow Convolutional Neural
  Network}. IEEE Access  \textbf{8},  66941--66950 (2020).
  \doi{10.1109/ACCESS.2020.2983182}

\bibitem{kant2020cwt}
Kant, P., Laskar, S.H., Hazarika, J., Mahamune, R.: {CWT Based Transfer
  Learning for Motor Imagery Classification for Brain computer Interfaces}. J.
  Neurosci. Methods  \textbf{345},  108886 (2020).
  \doi{10.1016/j.jneumeth.2020.108886}

\bibitem{kataoka2020pre}
Kataoka, H., Okayasu, K., Matsumoto, A., Yamagata, E., Yamada, R., Inoue, N.,
  Nakamura, A., Satoh, Y.: {Pre-training without Natural Images}. IJCV
  \textbf{130}(4),  990--1007 (2022). \doi{10.1007/s11263-021-01555-8}

\bibitem{kingma2014adam}
Kingma, D.P., Ba, J.: {Adam: A Method for Stochastic Optimization} (2015)

\bibitem{kostas2021bendr}
Kostas, D., Aroca-Ouellette, S., Rudzicz, F.: {BENDR: Using Transformers and a
  Contrastive Self-Supervised Learning Task to Learn From Massive Amounts of
  EEG Data}. Front. Hum. Neurosci.  \textbf{15},  653659 (2021).
  \doi{10.3389/fnhum.2021.653659}

\bibitem{lawhern2018eegnet}
Lawhern, V.J., Solon, A.J., Waytowich, N.R., Gordon, S.M., Hung, C.P., Lance,
  B.J.: {EEGNet: a compact convolutional neural network for EEG-based
  brain--computer interfaces}. J. Neural Eng.  \textbf{15}(5),  056013 (2018).
  \doi{10.1088/1741-2552/aace8c}

\bibitem{lopez2018brain}
L^^c3^^b3pez-Larraz, E., Sarasola-Sanz, A., Irastorza-Landa, N., Birbaumer, N.,
  Ramos-Murguialday, A.: {Brain-machine interfaces for rehabilitation in
  stroke: A review}. NeuroRehabilitation  \textbf{43}(1),  77--97 (2018).
  \doi{10.3233/nre-172394}

\bibitem{leeb2008bci}
Leeb, R., Brunner, C., M{\"u}ller-Putz, G., Schl{\"o}gl, A., Pfurtscheller, G.:
  {BCI Competition 2008--Graz data set B}. Graz University of Technology,
  Austria  \textbf{16}, ~1--6 (2008)

\bibitem{li2019novel}
Li, M.A., Han, J.F., Duan, L.J.: {A Novel MI-EEG Imaging With the Location
  Information of Electrodes}. IEEE Access  \textbf{8},  3197--3211 (2019).
  \doi{10.1109/ACCESS.2019.2962740}

\bibitem{li2016extracting}
Li, M.a., Luo, X.y., Yang, J.f.: {Extracting the nonlinear features of motor
  imagery EEG using parametric t-SNE}. Neurocomputing  \textbf{218},  371--381
  (2016). \doi{10.1016/j.neucom.2016.08.083}

\bibitem{liu2021densely}
Liu, T., Yang, D.: {A Densely Connected Multi-Branch 3D Convolutional Neural
  Network for Motor Imagery EEG Decoding}. Brain Sci.  \textbf{11}(2), ~197
  (2021). \doi{10.3390/brainsci11020197}

\bibitem{liu2021swin}
Liu, Z., Lin, Y., Cao, Y., Hu, H., Wei, Y., Zhang, Z., Lin, S., Guo, B.: {Swin
  Transformer: Hierarchical Vision Transformer using Shifted Windows}. In:
  ICCV. pp. 9992--10002 (2021). \doi{10.1109/ICCV48922.2021.00986}

\bibitem{luo2018exploring}
Luo, T.j., Zhou, C.l., Chao, F.: {Exploring spatial-frequency-sequential
  relationships for motor imagery classification with recurrent neural
  network}. BMC Bioinf.  \textbf{19}(1) (2018). \doi{10.1186/s12859-018-2365-1}

\bibitem{ma2021novel}
Ma, Q., Wang, M., Hu, L., Zhang, L., Hua, Z.: {A Novel Recurrent Neural Network
  to Classify EEG Signals for Customers' Decision-Making Behavior Prediction in
  Brand Extension Scenario}. Front. Hum. Neurosci.  \textbf{15},  610890
  (2021). \doi{10.3389/fnhum.2021.610890}

\bibitem{ma2019deep}
Ma, X., Qiu, S., Wei, W., Wang, S., He, H.: {Deep Channel-Correlation Network
  for Motor Imagery Decoding From the Same Limb}. IEEE Trans. Neural Syst.
  Rehabil. Eng.  \textbf{28}(1),  297--306 (2020).
  \doi{10.1109/TNSRE.2019.2953121}

\bibitem{van2008visualizing}
Van~der Maaten, L., Hinton, G.: {Visualizing Data using t-SNE}. JMLR
  \textbf{9}(86),  2579--2605 (2008)

\bibitem{mcinnes2018umap}
McInnes, L., Healy, J., Saul, N., Gro^^c3^^9fberger, L.: {UMAP: Uniform
  Manifold Approximation and Projection}. J. Open Source Softw.
  \textbf{3}(29), ~861 (2018). \doi{10.21105/joss.00861}

\bibitem{olivas2019classification}
Olivas-Padilla, B.E., Chacon-Murguia, M.I.: {Classification of multiple motor
  imagery using deep convolutional neural networks and spatial filters}. Appl.
  Soft Comput.  \textbf{75},  461--472 (2019). \doi{10.1016/j.asoc.2018.11.031}

\bibitem{planelles2014evaluating}
Planelles, D., Hortal, E., Costa, {\'A}., {\'U}beda, A., I{\'a}{\~n}ez, E.,
  Azor{\'\i}n, J.M.: {Evaluating Classifiers to Detect Arm Movement Intention
  from EEG Signals}. Sensors  \textbf{14}(10),  18172--18186 (2014).
  \doi{10.3390/s141018172}

\bibitem{roots2020fusion}
Roots, K., Muhammad, Y., Muhammad, N.: {Fusion Convolutional Neural Network for
  Cross-Subject EEG Motor Imagery Classification}. Computers  \textbf{9}(3),
  ~72 (2020). \doi{10.3390/computers9030072}

\bibitem{ILSVRC15}
Russakovsky, O., Deng, J., Su, H., Krause, J., Satheesh, S., Ma, S., Huang, Z.,
  Karpathy, A., Khosla, A., Bernstein, M., Berg, A.C., Fei-Fei, L.: {ImageNet
  Large Scale Visual Recognition Challenge}. IJCV  \textbf{115}(3),  211--252
  (2015). \doi{10.1007/s11263-015-0816-y}

\bibitem{bci2000}
Schalk, G., McFarland, D., Hinterberger, T., Birbaumer, N., Wolpaw, J.:
  {BCI2000: a general-purpose brain-computer interface (BCI) system}. IEEE
  Trans. Biomed. Eng.  \textbf{51}(6),  1034--1043 (2004).
  \doi{10.1109/TBME.2004.827072}

\bibitem{song2021transformer}
Song, Y., Jia, X., Yang, L., Xie, L.: {Transformer-based Spatial-Temporal
  Feature Learning for EEG Decoding}. arXiv preprint arXiv:2106.11170  (2021).
  \doi{10.48550/arXiv.2106.11170}

\bibitem{sun2021hybrid}
Sun, J., Cao, R., Zhou, M., Hussain, W., Wang, B., Xue, J., Xiang, J.: {A
  hybrid deep neural network for classification of schizophrenia using EEG
  Data}. Sci. Rep.  \textbf{11}(1), ~4706 (2021)

\bibitem{tabar2016novel}
Tabar, Y.R., Halici, U.: {A novel deep learning approach for classification of
  EEG motor imagery signals}. J. Neural Eng  \textbf{14}(1),  016003 (2016).
  \doi{10.1088/1741-2560/14/1/016003}

\bibitem{vaswani2017attention}
Vaswani, A., Shazeer, N., Parmar, N., Uszkoreit, J., Jones, L., Gomez, A.N.,
  Kaiser, L.u., Polosukhin, I.: {Attention is All you Need}. In: Guyon, I.,
  Luxburg, U.V., Bengio, S., Wallach, H., Fergus, R., Vishwanathan, S.,
  Garnett, R. (eds.) NeurIPS. vol.~30 (2017)

\bibitem{wairagkar2019modeling}
Wairagkar, M., Hayashi, Y., Nasuto, S.J.: {Modeling the Ongoing Dynamics of
  Short and Long-Range Temporal Correlations in Broadband EEG During Movement}.
  Front. Syst. Neurosci.  \textbf{13}, ~66 (2019).
  \doi{10.3389/fnsys.2019.00066}

\bibitem{wairagkar2021dynamics}
Wairagkar, M., Hayashi, Y., Nasuto, S.J.: {Dynamics of Long-Range Temporal
  Correlations in Broadband EEG During Different Motor Execution and Imagery
  Tasks}. Front. Neurosci.  \textbf{15} (2021). \doi{10.3389/fnins.2021.660032}

\bibitem{wang2018lstm}
Wang, P., Jiang, A., Liu, X., Shang, J., Zhang, L.: {LSTM-Based EEG
  Classification in Motor Imagery Tasks}. IEEE Trans. Neural Syst. Rehabil.
  Eng.  \textbf{26}(11),  2086--2095 (2018). \doi{10.1109/TNSRE.2018.2876129}

\bibitem{wang2019learning}
Wang, Q., Li, B., Xiao, T., Zhu, J., Li, C., Wong, D.F., Chao, L.S.: {Learning
  Deep Transformer Models for Machine Translation}. In: ACCL (2019).
  \doi{10.18653/v1/p19-1176}

\bibitem{wang2023internimage}
Wang, W., Dai, J., Chen, Z., Huang, Z., Li, Z., Zhu, X., Hu, X., Lu, T., Lu,
  L., Li, H., et~al.: {InternImage: Exploring Large-Scale Vision Foundation
  Models with Deformable Convolutions}. In: CVPR. pp. 14408--14419 (2023).
  \doi{10.1109/cvpr52729.2023.01385}

\bibitem{wang2018short}
Wang, Z., Cao, L., Zhang, Z., Gong, X., Sun, Y., Wang, H.: {Short time Fourier
  transformation and deep neural networks for motor imagery brain computer
  interface recognition}. Concurrency Comput. Pract. Exp.  \textbf{30}(23),
  e4413 (2018). \doi{10.1002/cpe.4413}

\bibitem{wei2023tc}
Wei, Y., Liu, Y., Li, C., Cheng, J., Song, R., Chen, X.: {TC-Net: A Transformer
  Capsule Network for EEG-based emotion recognition}. Comput. Biol. Med.
  \textbf{152},  106463 (2023). \doi{10.1016/j.compbiomed.2022.106463}

\bibitem{xie2022transformer}
Xie, J., Zhang, J., Sun, J., Ma, Z., Qin, L., Li, G., Zhou, H., Zhan, Y.: {A
  Transformer-Based Approach Combining Deep Learning Network and
  Spatial-Temporal Information for Raw EEG Classification}. IEEE Trans. Neural
  Syst. Rehabil. Eng.  \textbf{30},  2126--2136 (2022).
  \doi{10.1109/TNSRE.2022.3194600}

\bibitem{xu2018wavelet}
Xu, B., Zhang, L., Song, A., Wu, C., Li, W., Zhang, D., Xu, G., Li, H., Zeng,
  H.: {Wavelet Transform Time-Frequency Image and Convolutional Network-Based
  Motor Imagery EEG Classification}. IEEE Access  \textbf{7},  6084--6093
  (2018). \doi{10.1109/ACCESS.2018.2889093}

\bibitem{yang2018deep}
Yang, J., Yao, S., Wang, J.: {Deep Fusion Feature Learning Network for MI-EEG
  Classification}. IEEE Access  \textbf{6},  79050--79059 (2018).
  \doi{10.1109/ACCESS.2018.2877452}

\bibitem{yu2022metaformer}
Yu, W., Luo, M., Zhou, P., Si, C., Zhou, Y., Wang, X., Feng, J., Yan, S.:
  {MetaFormer is Actually What You Need for Vision}. In: CVPR. pp. 10819--10829
  (2022). \doi{10.1109/cvpr52688.2022.01055}

\bibitem{yu2016toward}
Yu, Y., Zhou, Z., Yin, E., Jiang, J., Tang, J., Liu, Y., Hu, D.: {Toward
  brain-actuated car applications: Self-paced control with a motor
  imagery-based brain-computer interface}. Comput. Biol. Med.  \textbf{77},
  148--155 (2016). \doi{10.1016/j.compbiomed.2016.08.010}

\bibitem{yun2019cutmix}
Yun, S., Han, D., Oh, S.J., Chun, S., Choe, J., Yoo, Y.: {CutMix:
  Regularization Strategy to Train Strong Classifiers with Localizable
  Features}. In: ICCV. pp. 6023--6032 (2019)

\bibitem{zhang2017mixup}
Zhang, H., Cisse, M., Dauphin, Y.N., Lopez-Paz, D.: {mixup: Beyond Empirical
  Risk Minimization}. In: ICLR (2017)

\bibitem{zhang2018mindid}
Zhang, X., Yao, L., Kanhere, S.S., Liu, Y., Gu, T., Chen, K.: {MindID: Person
  Identification from Brain Waves through Attention-based Recurrent Neural
  Network}. IMWUT  \textbf{2}(3),  1--23 (2018). \doi{10.1145/3264959}

\bibitem{zhao2020deep}
Zhao, X., Zhao, J., Liu, C., Cai, W.: {Deep Neural Network with Joint
  Distribution Matching for Cross-Subject Motor Imagery Brain-Computer
  Interfaces}. BioMed Res. Int.  \textbf{2020},  1--15 (2020).
  \doi{10.1155/2020/7285057}

\bibitem{zhao2019multi}
Zhao, X., Zhang, H., Zhu, G., You, F., Kuang, S., Sun, L.: {A Multi-Branch 3D
  Convolutional Neural Network for EEG-Based Motor Imagery Classification}.
  IEEE Trans. Neural Syst. Rehabil. Eng.  \textbf{27}(10),  2164--2177 (2019).
  \doi{10.1109/TNSRE.2019.2938295}

\bibitem{zhu2019study}
Zhu, K., Wang, S., Zheng, D., Dai, M.: {Study on the effect of different
  electrode channel combinations of motor imagery EEG signals on classification
  accuracy}. J. Eng.  \textbf{2019}(23),  8641--8645 (2019).
  \doi{10.1049/joe.2018.9073}

\end{thebibliography}
\end{document}